%% file: main.tex
\documentclass[10pt,twocolumn,letterpaper]{article}

\usepackage{cvpr}
\usepackage{times}
\usepackage{epsfig}
\usepackage{graphicx}
\usepackage{amsmath}
\usepackage{amssymb}

\usepackage{ifthen}
\newboolean{ack}
\setboolean{ack}{true}

\usepackage{booktabs}       
\usepackage{amsfonts}       
\usepackage{nicefrac}       
\usepackage{microtype}      
\usepackage{multirow}
\usepackage{subcaption}
\usepackage{algpseudocode}
\usepackage{blindtext}
\usepackage{xfrac}
\usepackage{bm}
\usepackage{chngcntr}

\usepackage{array}
\usepackage{placeins}
\usepackage{rotating}
\usepackage{multicol}
\usepackage{multirow}
\usepackage{enumitem}

\usepackage{color}
\usepackage{fancyhdr,graphicx,amsmath,amssymb}
\usepackage[ruled,vlined]{algorithm2e}
\usepackage{longtable}

\usepackage[pagebackref=true,breaklinks=true,letterpaper=true,colorlinks,bookmarks=false]{hyperref}

\cvprfinalcopy 

\ifcvprfinal\fi
\begin{document}

\title{Lifelong Machine Learning with\\Deep Streaming Linear Discriminant Analysis}

\author{Tyler L. Hayes$^1$ \qquad \qquad Christopher Kanan$^{1,2,3}$\\
$^1$Rochester Institute of Technology\qquad $^2$Paige \qquad $^3$Cornell Tech\\
  \texttt{tlh6792@rit.edu, kanan@rit.edu} \\
}

\maketitle

\begin{abstract}
    When an agent acquires new information, ideally it would immediately be capable of using that information to understand its environment. This is not possible using conventional deep neural networks, which suffer from catastrophic forgetting when they are incrementally updated, with new knowledge overwriting established representations. A variety of approaches have been developed that attempt to mitigate catastrophic forgetting in the incremental batch learning scenario, where a model learns from a series of large collections of labeled samples. However, in this setting, inference is only possible after a batch has been accumulated, which prohibits many applications. An alternative paradigm is online learning in a single pass through the training dataset on a resource constrained budget, which is known as streaming learning. Streaming learning has been much less studied in the deep learning community. In streaming learning, an agent learns instances one-by-one and can be tested at any time, rather than only after learning a large batch. Here, we revisit streaming linear discriminant analysis, which has been widely used in the data mining research community. By combining streaming linear discriminant analysis with deep learning, we are able to outperform both incremental batch learning and streaming learning algorithms on both ImageNet ILSVRC-2012 and CORe50, a dataset that involves learning to classify from temporally ordered samples\footnote{\url{https://github.com/tyler-hayes/Deep_SLDA}}. 
\end{abstract}


\section{Introduction}
\label{sec:intro}

For many real-time applications, an agent must be capable of immediate online learning of each training instance, without the ability to loop through the entire dataset and while being subject to severe resource constraints. The ability to do online learning under these constraints from non-stationary data streams in a single pass is known as \emph{streaming learning}~\cite{aggarwal2004demand,bifet2009adaptive,domingos2000mining,gaber2007survey,gama2013evaluating,hayes2019memory,hulten2001mining,read2012scalable}. This training paradigm presents unique challenges to agents including limited access to computational resources in terms of memory and compute time and inference must be able to be performed at any time during training~\cite{gama2010knowledge}.

Deep neural networks (DNNs) are the dominant approach in computer vision for inferring semantic information from sensors such as cameras, but conventional DNNs are not capable of being incrementally updated or learning quickly from individual instances. Incrementally updating a DNN is challenging due to the stability-plasticity dilemma~\cite{abraham2005memory}. To learn, a DNN must alter its weights, but altering weights that are critical for retaining past knowledge can cause forgetting. When a DNN is incrementally updated with a temporal stream of data that is not independent and identically distributed (iid), this dilemma typically manifests as catastrophic forgetting~\cite{mccloskey1989}. Rather than gradually losing the ability to work well on past information, catastrophic forgetting refers to how learning only a small amount of new information can cause the complete loss of ability to operate on previously learned tasks. 

\begin{figure}[t]
    \centering
    \includegraphics[width=0.45\textwidth]{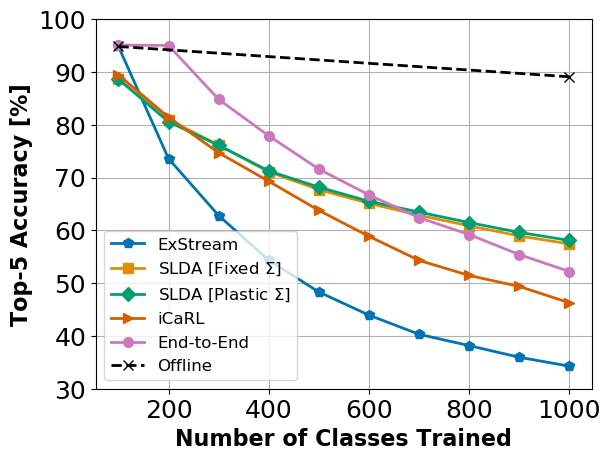}
    \caption{Learning curve for incremental ImageNet. Our Deep SLDA approach achieves the best final top-5 accuracy, while running over 100 times faster and using 1,000 times less memory than the iCaRL and End-to-End models.}
    \label{fig:main-results-imagenet}
    \vspace{-4mm}
\end{figure}

\begin{figure*}[th!]
    \centering
    \begin{subfigure}[b]{0.46\textwidth}
        \centering
        \includegraphics[width=\textwidth]{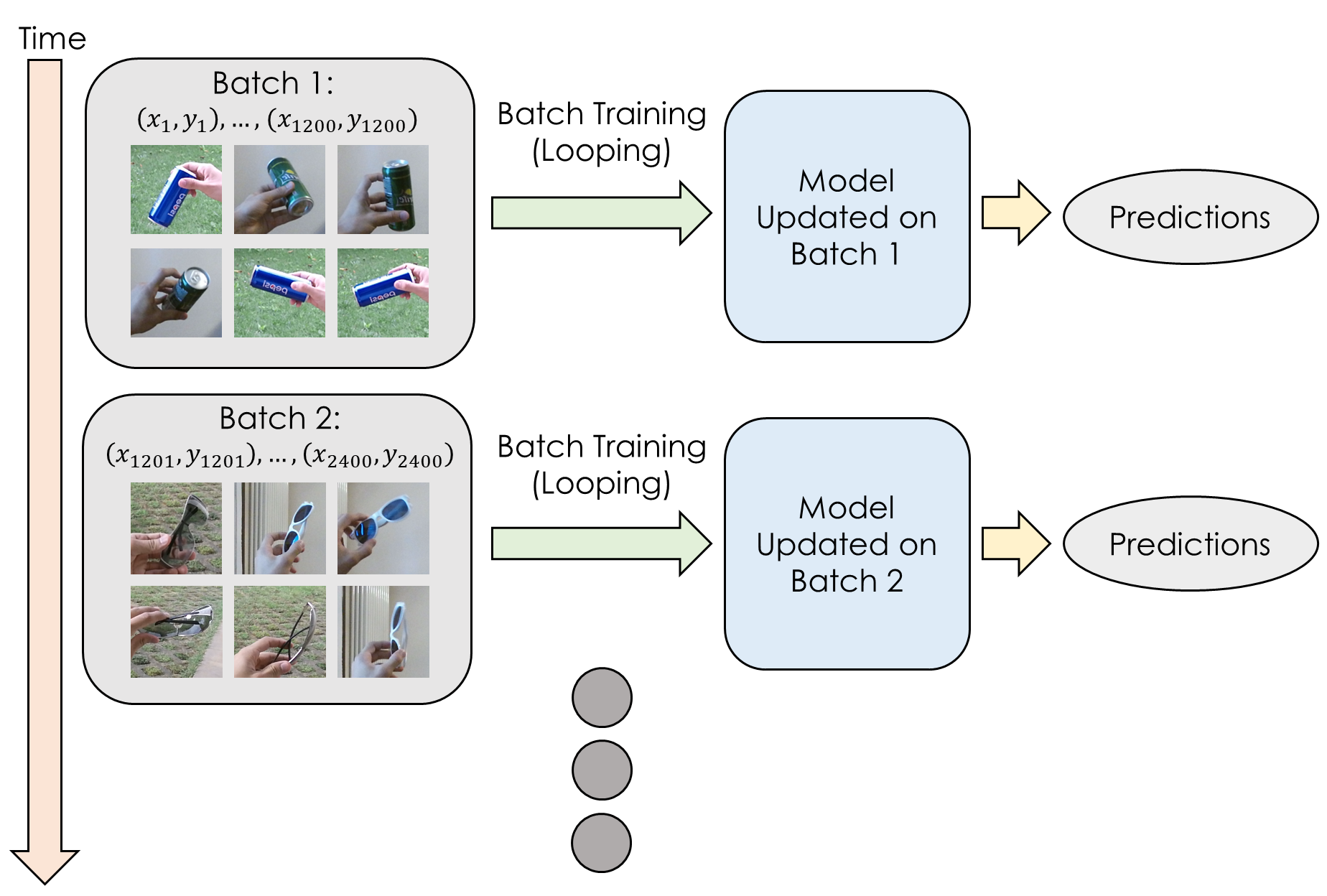}
        \caption{Incremental Batch Learning}
        \label{fig:paradigm-continual}
    \end{subfigure} 
    \begin{subfigure}[b]{0.46\textwidth}
        \centering
        \includegraphics[width=0.78\textwidth]{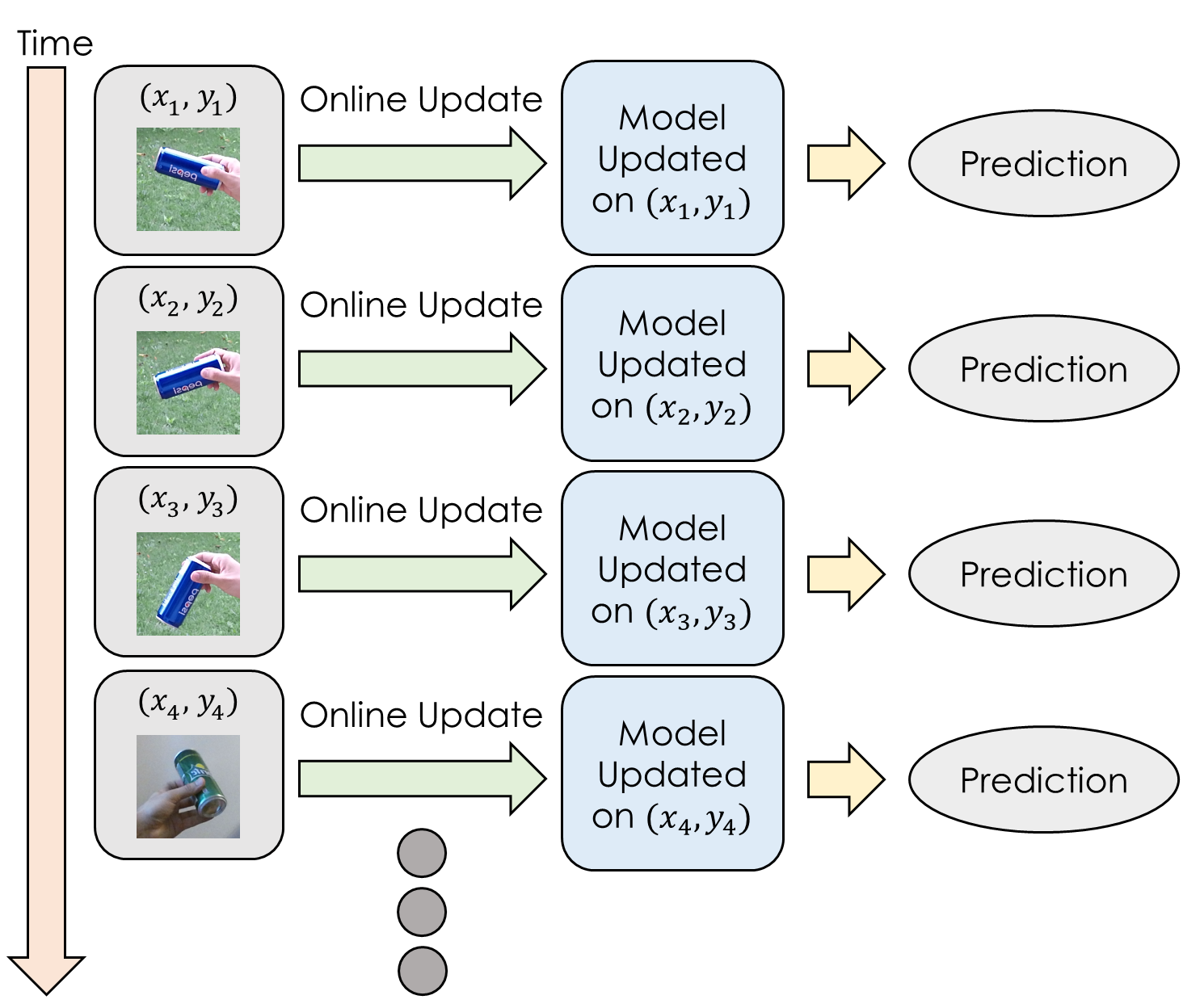}
        \caption{Streaming Learning}
        \label{fig:paradigm-streaming}
    \end{subfigure}
    \caption{Streaming learning requires agents to learn sample-by-sample in real time, making it better suited for embedded applications than incremental batch learning. For example, in our experiments on CORe50, each incremental batch consists of 1,200 samples (2 classes with 600 samples each). For ImageNet, each incremental batch consists of $\sim$13,000 images (100 classes with $\sim$1,300 samples each). All examples must be seen by the model multiple times before inference can be performed. In contrast, for streaming learning, new information can be learned and used immediately. Images are from CORe50.
    }
    \label{fig:experimental-paradigms}
    \vspace{-3mm}
\end{figure*}

In the past few years, much effort has been directed at creating modified DNNs that can be incrementally updated without catastrophic forgetting. The vast majority of these systems operate in the incremental batch learning framework~\cite{castro2018end,chaudhry2018riemannian,fernando2017,kemker2018fearnet,kemker2018forgetting,rebuffi2016icarl,zenke2017continual}. In this setting, the DNN receives a series of large batches of new labeled samples. After a batch has been received, the DNN loops over the batch until it is adequately learned, and then the DNN can be tested on information in that batch and previous batches. Most incremental batch learning methods utilize partial rehearsal or pseudo-rehearsal~\cite{kemker2018forgetting}. Partial rehearsal involves storing some examples from each batch in auxiliary memory and then mixing them with the current batch being learned. Instead of storing examples, pseudo-rehearsal uses a generative model to create examples from earlier batches. While pseudo-rehearsal seems appealing, the generator often has just as many, if not more, parameters than the DNN used for inference and continuously generating examples to learn is computationally expensive. Neither approach is ideal for a model with limited resources for fast on-device learning. Cloud computing can avoid this problem, but it can lead to privacy, security, and latency issues.

Streaming learning has been little studied with DNNs. Numerous streaming classifiers have been explored in the data mining community, but these methods have primarily been assessed with low-dimensional data streams and most are slow to train~\cite{gaber2007survey}. Here, we explore the use of deep Streaming Linear Discriminant Analysis (SLDA)~\cite{pang2005incremental} for training the output layer of a convolutional neural network (CNN) incrementally, which has not been done before. We validate performance on large-scale image classification datasets under multiple data orderings that cause catastrophic forgetting in conventional DNNs. Since SLDA only trains the output layer of a CNN and does not store any previous data, it is a lightweight classifier that can be easily deployed on embedded platforms.

\textbf{This paper makes the following contributions:}
\begin{enumerate}[noitemsep, nolistsep]
    \item We describe the deep SLDA algorithm. We are the first to use SLDA for the classification of features from a deep CNN on large-scale image classification datasets. 
    \item We demonstrate that deep SLDA can surpass state-of-the-art streaming learning algorithms.
    \item Using both incremental ImageNet ILSVRC-2012 and CORe50, we demonstrate that deep SLDA can exceed recent methods for doing incremental batch learning, which is an easier problem, even though these methods update their hidden layers. Compared to these methods deep SLDA is over 100 times faster to train and uses 1,000 times less memory. 
\end{enumerate}


\section{Problem Formulations \& Related Work}
\label{sec:related-work}

\subsection{Streaming Learning}

In incremental batch learning, an agent learns a dataset $\mathcal{D}$ that is broken up into $T$ distinct batches $B_t$, each of size $N_t$. At time $t$, it only has access to $B_t$, but it may loop over $B_t$ multiple times. Testing occurs between batches. Conversely, in \emph{streaming learning}, an agent learns examples one at a time ($N_t=1$) in a single pass through the dataset (see Fig.~\ref{fig:experimental-paradigms}). Mirroring animal learning, the agent can be evaluated at any point and it cannot loop over any portion of the dataset. In our setup, we assume the agent is learning to classify an input with no contextual information about the task.

While much progress has been made in mitigating catastrophic forgetting for neural networks in the incremental batch learning paradigm~\cite{castro2018end,chaudhry2018riemannian,fernando2017,kemker2018fearnet,kemker2018forgetting,rebuffi2016icarl,zenke2017continual}, there is still a large gap between incremental batch learners and offline models~\cite{kemker2018forgetting}, and much less progress has been made in the streaming paradigm~\cite{hayes2019memory}.

\subsection{Methods for Incremental Batch Learning}

Multiple approaches have been explored for mitigating forgetting, including regularizing weights to remain close to their previous values~\cite{aljundi2018memory,chaudhry2018riemannian,dhar2019learning,kirkpatrick2017,ritter2018online,serra2018overcoming,zenke2017continual}, promoting sparse weight updates to mitigate interference~\cite{FEL}, and ensembling multiple classifiers~\cite{fernando2017}, but recently, models that incorporate rehearsal (i.e., replay) have demonstrated the most success~\cite{castro2018end,hayes2019memory,kemker2018fearnet,kemker2018forgetting,lee2019overcoming,ostapenko2019learning,rebuffi2016icarl}. Rehearsal can come in the form of partial rehearsal where an agent maintains a subset of previous examples that are mixed with new samples to update the network. Partial rehearsal has been widely adopted by methods such as iCaRL~\cite{rebuffi2016icarl} and End-to-End Incremental learning~\cite{castro2018end}. In conjunction with storing and replaying previous samples, many methods use a distillation loss~\cite{hinton2015distilling} to regularize weight updates so that the network does not drift far from its previous solution~\cite{castro2018end,hou2018lifelong,javed2018revisiting,rebuffi2016icarl,wu2018memory}. 

Instead of storing examples explicitly, pseudo-rehearsal methods learn to model the distribution of previous training samples and generate `pseudo-examples' to mix with new data during updates using a generative model such as an auto-encoder~\cite{kemker2018fearnet}. While rehearsal methods have demonstrated success and are widely used, they are memory intensive (i.e., storing explicit past samples in the case of rehearsal and storing an encoder and decoder for pseudo-rehearsal) and each incremental update requires more compute time due to the large number of samples. Additionally, generative models such as auto-encoders and generative adversarial networks can often be slow and difficult to train.

Although there has been much recent interest in incremental batch learning~\cite{castro2018end,chaudhry2018riemannian,kemker2018fearnet,kemker2018forgetting,rebuffi2016icarl,zenke2017continual}, this setting is not appropriate for models deployed in real-time environments. Waiting for a batch of information to accumulate before inference possibly restricts many applications.

\subsection{Methods for Streaming Learning}

Streaming learning has been studied since at least 1980~\cite{munro1980selection}, and many popular streaming classifiers come from the data mining community.  Hoeffding Decision Trees~\cite{bifet2009adaptive,domingos2000mining,gama2006decision,hulten2001mining,ikonomovska2011learning} incrementally grow decision trees over time under the Hoeffding bound theoretical guarantees. Another widely used method is ensembling multiple classifiers~\cite{bifet2009new,brzezinski2014reacting,wang2003mining}. However, both Hoeffding Decision Trees and ensemble methods are slow to train~\cite{gaber2007survey}, making them ill-suited choices for many embedded applications operating in real-time. There have been shallow neural networks designed for streaming learning, including ARTMAP networks~\cite{carpenter1992fuzzy,carpenter1991artmap,williamson1996gaussian}; however, ARTMAP is sensitive to the order in which training data is presented and it is not capable of representation learning.

Recently, there have been two notable attempts to marry streaming learning with DNNs: 1) the gradient episodic memory (GEM) family of algorithms~\cite{chaudhry2019efficient,lopez2017gradient} and 2) ExStream~\cite{hayes2019memory}. The GEM family of models use regularization to constrain weight updates on new tasks such that the loss incurred on previously stored training samples can decrease, but not increase. While popular, they cannot readily be used for embedded applications because they require the task label during inference. If task labels are not provided to these models during testing, model performance will significantly degrade, deeming the models unusable~\cite{chaudhry2018riemannian,farquhar2018towards,kemker2018forgetting}.

The second method for updating a DNN in the streaming setting is the ExStream algorithm~\cite{hayes2019memory}.  Similar to deep SLDA, ExStream can update only the fully-connected layers of a CNN. ExStream uses partial rehearsal to combat forgetting by maintaining a buffer of prototypes for each class. When it receives a new instance to learn, it stores that example in its associated class-specific buffer and then, if the buffer is full, it merges the two closest exemplars in its buffer. The entire buffer is then used to update the fully-connected layers with a single iteration of stochastic gradient descent. While it is one of the only deep streaming classifiers, ExStream still has bottlenecks in terms of memory and compute due to its rehearsal mechanisms.

Especially relevant to this paper are SLDA~\cite{pang2005incremental} and Streaming Quadratic Discriminant Analysis (SQDA). SLDA maintains one running mean per class and a shared covariance matrix that can be held fixed, or updated using an online update. To make predictions, SLDA assigns the label to an input of the closest Gaussian computed using the running class means and covariance matrix. Similar to SLDA, SQDA assumes that each class is normally distributed. However, instead of assuming each class has the same covariance, SQDA assumes each class has its own covariance, which can be updated using online estimates. Due to the maintenance of one covariance matrix per class, SQDA requires more memory and compute resources as compared to SLDA, making it less suitable for on-device learning. For example, using SQDA with embeddings from a ResNet-18~\cite{He_2016_CVPR} architecture on a 1,000 class dataset such as ImageNet would require storing 1,000 covariance matrices of dimension 512$\times$512, whereas SLDA would only require storing a single 512$\times$512 covariance matrix. Further, it was shown in \cite{lee2018simple} that the estimated posterior distribution of LDA is equivalent to the softmax classifier often used with modern neural networks, thus motivating the use of SLDA.

\section{Deep Streaming LDA}
\label{sec:deep-slda}

Formally, we incrementally train a CNN $y_t = F\left( G \left( \mathbf{X}_t \right) \right)$ in a streaming manner, where $\mathbf{X}_t$ is the input image and $y_t$ is the output category. We decompose the network into two nested functions:  $G\left( \cdot \right)$ consists of the first $J$ layers of the CNN (with parameters $\theta_G$) and $F\left( \cdot \right)$ consists of the last fully-connected layer (with parameters $\theta_F$). We assume the output of $G\left( \cdot \right)$ is a vector, which could be produced by pooling across spatial locations of a feature map, flattening the feature map, etc. Because the filters learned in the early layers of a CNN vary little across large natural image datasets and are highly transferable~\cite{yosinski2014transferable}, SLDA keeps $\theta_G$ fixed, and focuses on training $F\left( \cdot \right)$ in a streaming manner. We discuss how $G\left( \cdot \right)$ is trained during a \emph{base initialization} phase in  Sec.~\ref{sec:base-initialization}, which is common in recent incremental batch learning literature~\cite{castro2018end,rebuffi2016icarl}.

SLDA is an online extension of LDA. It is used in the data mining community to perform streaming learning from low-dimensional data streams. We adapt SLDA to train a linear decoder $F(\cdot)$ for $G(\cdot)$, i.e.,
\begin{equation}
\label{eq:pred}
    F(G(\mathbf{X}_t)) = \mathbf{W} \mathbf{z}_t + \mathbf{b} \enspace ,
\end{equation}
where $\mathbf{z}_t = G\left( \mathbf{X}_t \right) \in \mathbb{R}^{d}$ is a vector, $K$ is the total number of categories and $d$ is the dimensionality of the data with both weight matrix $\mathbf{W} \in \mathbb{R}^{K \times d}$ and bias vector $\mathbf{b} \in \mathbb{R}^K$ being updated online.

SLDA stores one mean vector per class $\bm{\mu}_k \in \mathbb{R}^{d}$ with an associated count $c_{k} \in \mathbb{R}$ and a single shared covariance matrix $\mathbf{\Sigma} \in \mathbb{R}^{d \times d}$. When a new datapoint $\left(\mathbf{z}_t,y\right)$ arrives, the mean vector and associated counter are updated as:
\begin{equation}
\label{eq:pos-mean}
        \bm{\mu}_{(k=y,t+1)} \gets  \frac{c_{(k=y,t)}\bm{\mu}_{(k=y,t)} + \mathbf{z}_t}{c_{(k=y,t)}+1}
\end{equation}    
\begin{equation}
\label{eq:pos-count}
    c_{(k=y,t+1)} = c_{(k=y,t)} + 1 \enspace ,
\end{equation}
where $\bm{\mu}_{(k=y,t)}$ is the mean vector for class $y$ at time $t$ and $c_{(k=y,t)}$ is the associated $y$-th counter.

We use shrinkage regularization to compute the precision matrix, i.e., $\mathbf{\Lambda} = \left[\left(1-\varepsilon\right)\mathbf{\Sigma} + \varepsilon \mathbf{I}\right]^{-1}$, where $\varepsilon=10^{-4}$ is the shrinkage parameter and $\mathbf{I} \in \mathbb{R}^{d \times d}$ is the identity matrix. We explore two SLDA variants: 1) using a frozen covariance matrix after base initialization (see Sec.~\ref{sec:base-initialization}), and 2) streaming updates of the covariance matrix. With a frozen covariance matrix, its inverse is computed only once, but updating it requires the inverse to be computed before inference.

For the SLDA variant that updates the covariance matrix online, we use the update from \cite{dasgupta2007line}, i.e.,
\begin{equation}
    \mathbf{\Sigma}_{t+1} = \frac{t\mathbf{\Sigma}_{t}+\Delta_{t}}{t+1} \enspace ,
\end{equation}
where $\Delta_t$ is computed as:
\begin{equation}
    \Delta_{t} = \frac{t\left(\mathbf{z}_{t}-\bm{\mu}_{(k=y,t)}\right)\left(\mathbf{z}_{t}-\bm{\mu}_{(k=y,t)}\right)^{T}}{t+1} \enspace .
\end{equation}

To compute predictions, we use Eq.~\ref{eq:pred} and compute $\mathbf{w}_k$, i.e., the rows of $\mathbf{W}$, as:
\begin{equation}
    \mathbf{w}_k = \mathbf{\Lambda} \bm{\mu}_{k} 
\end{equation}
and $b_k$, i.e., the individual elements of $\mathbf{b}$, as:
\begin{equation}
    b_k = - \frac{1}{2}\left(\bm{\mu}_{k} \cdot \mathbf{\Lambda} \bm{\mu}_{k}\right) \enspace ,
\end{equation}
where $\cdot$ denotes the dot product.

SLDA is resistant to catastrophic forgetting because its running means for each class are  independent, which directly avoids the stability-plasticity dilemma. While the covariance matrix can change over time and is sensitive to class ordering, the changes to it result in, at most, gradual forgetting.

\section{Experiments}
\label{sec:experiments}

\subsection{Baseline \& Comparison Models}
\label{sec:comparison-models}


We compare SLDA with several baselines under two paradigms, i.e., those that don't require task labels and those that do. Since SLDA does not require task labels, our main baselines consist of models that operate in this paradigm. While many recent incremental batch learning methods perform multiple loops over a data batch, SLDA is a streaming method that learns per instance. Despite this advantage for incremental batch learners, we compare SLDA against several recent methods. For all experiments we incrementally train the ResNet-18~\cite{He_2016_CVPR} CNN architecture. We assess these streaming and incremental batch learning methods:
\begin{itemize}[noitemsep, nolistsep]
    \item \textbf{Deep SLDA} -- We compare two versions of deep SLDA for updating the classification layer of a CNN. One version uses a covariance matrix that is computed during base initialization (see Sec.~\ref{sec:base-initialization}) and then kept fixed. The other version uses a covariance matrix that is incrementally updated during streaming learning. 
    \item \textbf{Fine-Tuning} -- This is a streaming baseline where the CNN is fine-tuned sample-by-sample with a single epoch through the dataset. No buffer is used, and this approach suffers from catastrophic forgetting~\cite{kemker2018forgetting}. We compare two  versions: 1) update only the output layer ($\theta_F$); and 2) update the entire network ($\theta_F$ and $\theta_G$). 
    \item \textbf{ExStream} -- Like SLDA, ExStream is a streaming learning algorithm; however, it can only update fully-connected layers of the network. It maintains prototype buffers by storing an incoming vector and merging the two closest vectors in the buffer~\cite{hayes2019memory}. After the buffer is updated, its contents are used to train fully-connected layers in a network. We use ExStream to train the output layer of the network.
    It achieved state-of-the-art performance on the CORe50~\cite{lomonaco2017core50} streaming dataset.
    \item \textbf{iCaRL} -- iCaRL~\cite{rebuffi2016icarl} is a popular incremental batch learning method designed for incremental class learning, where each batch must contain two or more categories, and these classes are not seen in later batches. Without significant changes, it cannot operate in other ordering scenarios. To mitigate catastrophic forgetting, iCaRL stores raw images from earlier batches for partial rehearsal and uses distillation with these stored examples to prevent weights from drifting too far from their previous values. To make predictions, iCaRL uses the Nearest Class Mean classifier in feature space. iCaRL updates the entire CNN ($\theta_F$ and $\theta_G$).
    \item \textbf{End-to-End} -- The End-to-End Incremental Learning model~\cite{castro2018end} is an incremental batch learning method that is a modification of iCaRL. It achieved state-of-the-art performance on incremental class learning with ImageNet. Rather than a Nearest Class Mean classifier, it uses the CNN's output layer. It uses several augmentation strategies to get more out of its buffer, including brightness enhancements, contrast normalization, random crops, and mirror flips. It has the same limitations as iCaRL: it cannot do streaming learning and can only operate in the incremental class batch learning setting.
    \item \textbf{Offline} -- Offline is a model that is trained in an offline, non-streaming manner. It is used to normalize performance and it serves as an upper bound on an incremental learner's performance. We compare two versions: 1) update $\theta_F$ only, and 2) update $\theta_F$ and $\theta_G$. 
\end{itemize}
All models use the same offline base CNN initialization procedure (see Sec.~\ref{sec:base-initialization}). Subsequently, ExStream and SLDA re-start a streaming learning phase from the beginning of the dataset to train the output layer while keeping remaining parameters fixed. With the exception of SLDA,  we use cross-entropy classification loss and stochastic gradient descent with momentum to train the CNN. End-to-End and iCaRL additionally use distillation for targets. While it would be interesting to develop a deep SQDA method, maintaining a full covariance matrix for each class is not feasible for high-dimensional, many-class scenarios, like ImageNet.

\subsubsection{Baselines Requiring Task Labels}

As mentioned earlier, the GEM models~\cite{chaudhry2019efficient,lopez2017gradient} require a task label to be provided at test time, which is not compatible with our main experimental setup, and these labels are typically not available in embedded applications where streaming learning would be most useful. Regardless, we provide a small-scale experiment to compare SLDA against a newer variant of GEM, Averaged GEM (A-GEM)~\cite{chaudhry2019efficient}, and several other recent regularization models in Sec.~\ref{sec:core50-results}.


\subsection{Datasets, Data Orderings, \& Evaluation}
\label{sec:datasets-orderings-eval}

We compare the models on the ImageNet ILSVRC-2012~\cite{russakovsky2015imagenet} and CORe50~\cite{lomonaco2017core50} datasets. ImageNet has over one million images from the internet of 1,000 object categories.  Following others~\cite{castro2018end,rebuffi2016icarl}, all ImageNet models start from an offline base initialization of 100 randomly selected classes, and then performance is computed every 100 classes on all classes learned. We use top-5 accuracy for ImageNet.

Although ImageNet contains many categories, it is not ideal for streaming learning because it does not have temporally ordered video frames, which more closely models animal perception. To address this, we use the CORe50 dataset. It contains short 15 second video sequences of an object moving. It has 10 object categories, each with 5 distinct objects, that were recorded under 11 different environmental conditions (e.g., various backgrounds, outdoors/indoors, etc.). The videos were originally recorded at 20 fps, but we sample them at 1 fps and use the 128$\times$128 bounding box crops. Due to the smoothness of the videos, down-sampling the video frame rate is a common practice with CORe50~\cite{hayes2019memory,parisi2018lifelong}. We use the train/test split suggested in \cite{lomonaco2017core50}, which results in 600 train and 225 test images per class. Since CORe50 consists of temporally ordered video sequences, the order in which the data are presented will affect the final results. For this reason, we explore four different orderings of the dataset as proposed in \cite{hayes2019memory}: 1) iid where all frames are randomly shuffled, 2) class iid where all of the frames are shuffled within each class, 3) instance where videos are temporally organized by object instances, and 4) class instance where videos are temporally organized by object instances by class. We evaluate each method on all test data every 1,200 samples and report metrics in terms of top-1 accuracy. For CORe50, we run each experiment with 10 different permutations of each ordering and report the mean performance across all runs.

Following \cite{hayes2019memory,kemker2018forgetting}, we measure performance using the normalized metric:
\begin{equation}
    \Omega_{\mathrm{all}} = \frac{1}{T}\sum_{t=1}^{T} \frac{\alpha_{t}}{\alpha_{\mathrm{offline},t}} \enspace ,
\end{equation}
where $\alpha_{t}$ is the incremental learner's performance at time $t$ and $\alpha_{\mathrm{offline},t}$ is the optimized offline performance trained on all data until time $t$. This approach assumes the offline model is an approximate upper bound. $\Omega_{\mathrm{all}}$ makes it easier to compare performance across datasets and orderings. Usually $\Omega_{\mathrm{all}}$ is in the range $[0,1]$, but if an incremental learner outperforms the offline baseline, it is possible for $\Omega_{\mathrm{all}}> 1$.


\input{tables/main-results-table.tex}

\subsection{Network Initialization}
\label{sec:base-initialization}

For the ImageNet experiments, we follow others and initialize $F(\cdot)$ and $G(\cdot)$ for each model with 100 fixed, but randomly selected classes~\cite{castro2018end,rebuffi2016icarl}. Note that $F(\cdot)$ and $G(\cdot)$ are only initialized on 100 classes from ImageNet and the remaining 900 classes are learned incrementally. For CORe50, we first initialize $F(\cdot)$ and $G(\cdot)$ with pre-trained ImageNet weights. We then replace the last fully-connected layer with a layer containing only 10 output units, and fine-tune $F(\cdot)$ and $G(\cdot)$ on 1,200 samples from CORe50, where the 1,200 selected samples are dependent on the data ordering, but fixed across models. Based on the subset of CORe50 that we use, each class consists of exactly 600 training samples, so for the class iid and class instance orderings, 1,200 samples corresponds to exactly 2 classes. Note that we use the same base initialization phase for all models on both ImageNet and CORe50 for fair comparison. For SLDA, we initialize the covariance matrix on this same base initialization data using the Oracle Approximating Shrinkage estimator~\cite{chen2010shrinkage}.


\subsection{Main Results}
\label{sec:main-results}

For ImageNet, we follow current incremental batch learning models~\cite{castro2018end,rebuffi2016icarl} and report top-5 accuracy after every 100 classes are learned on all previous classes. We use the pre-trained ResNet-18 model from PyTorch as our final offline accuracy for normalizing $\Omega_{\mathrm{all}}$, which achieves 89.08\% top-5 accuracy. For End-to-End, we use numbers provided by the authors for ImageNet and do not include results for CORe50 since we were not able to run the model ourselves.

For CORe50, we evaluate each model after every 1,200 samples are observed. For CORe50, we report top-1 accuracy and normalize $\Omega_{\mathrm{all}}$ to the offline learner, which achieves 93.62\% accuracy at the final time-step of the iid ordering. The iCaRL and End-to-End incremental batch learning models are trained on batches of 100 classes at a time for ImageNet and two classes at a time for CORe50, where they may loop over the batches until they have learned them. This gives these models a significant advantage over the SLDA and ExStream streaming models. Parameter settings for all models are in the Appendix. We report our final $\Omega_{\mathrm{all}}$ scores in Table~\ref{tab:classification-results} and a forgetting curve for ImageNet is in Fig.~\ref{fig:main-results-imagenet}.

\subsubsection{ImageNet Results}

Although SLDA cannot train the CNN's hidden layers, it outperforms iCaRL overall and ends with a higher accuracy than End-to-End on ImageNet (see Fig.~\ref{fig:main-results-imagenet}). Updating the SLDA covariance matrix only yielded marginal improvement ($\sim$0.5\%). This is likely a result of the covariance matrix having good feature representations from being pre-initialized on 100 classes of ImageNet. The streaming models without a replay buffer suffer from catastrophic forgetting and achieve poor overall performance.

\subsubsection{CORe50 Results}
\label{sec:core50-results}

Results on CORe50 resemble those on ImageNet. SLDA with a plastic covariance matrix outperforms SLDA with a fixed covariance matrix, ExStream, iCaRL, and the streaming model without a replay buffer. While updating the covariance matrix for SLDA on ImageNet only yielded a small improvement, for CORe50 it resulted in a large boost in performance across all four orderings. This is likely due to the base initialization for ImageNet having 100 classes, whereas the base initialization for CORe50 only had 1,200 samples, meaning the initial covariance matrix was not representative of the entire training set. Remarkably, SLDA with a plastic covariance matrix performed almost as well as the full offline learner for the iid ordering, and was close to the offline performance for the other three orderings. SLDA with a plastic covariance performed on par with the offline model that only trained the output layer, demonstrating its robustness to various orderings. The streaming model without a replay buffer performed well for the iid and instance orderings where classes are revisited, but performed poorly for the orderings where classes were visited only once. Although iCaRL is a top performer for ImageNet, ExStream and both variants of SLDA performed better on the class iid and class instance orderings of CORe50.

In our setup, we assume the agent is learning to model $P(C=k | \mathbf{X})$, where $k \in C$ is a class label and $\mathbf{X}$ is an input; however, some algorithms learn to model $P(C=k | \mathbf{X}, i)$ where $i$ is the task label that must be provided with the input. We compare SLDA against several regularization methods that require task labels and use constraints to ensure that parameters do not change too much from their previous values during incremental training. Namely, we compare against the Synaptic Intelligence (SI)~\cite{zenke2017continual}, Elastic Weight Consolidation (EWC)~\cite{kirkpatrick2017}, Memory Aware Synapses (MAS)~\cite{aljundi2018memory}, Riemannian Walk for Incremental Learning (RWALK)~\cite{chaudhry2018riemannian}, and A-GEM~\cite{chaudhry2019efficient} approaches both with and without task labels during inference on both class orderings of CORe50 in Table~\ref{table:regularization-results}. To obtain results for SLDA and Offline with task labels, we mask off probabilities pertaining to classes not seen within a particular batch of data. Since our experiments with CORe50 require models to learn batches of two classes at a time, providing task labels during test time reduces the problem to binary classification, which is much easier than the 10-way classification problem without task labels. Regardless, SLDA outperforms all regularization methods both with and without task labels, even when the covariance matrix $\Sigma$ is held fixed, further demonstrating its robustness.

\input{tables/regularization-table.tex}

\subsection{Additional Experiments and Analysis}
\label{sec:additional-experiments}

\paragraph{Compute.}
SLDA outperforms ExStream and iCaRL by a large margin, while running in significantly less time. For example, ExStream requires 31 hours to run on ImageNet and iCaRL requires 62 hours, while SLDA with a plastic covariance matrix only requires 30 minutes on the same hardware. Less compute is desirable for embedded agents that must quickly learn and adapt to new information.

\paragraph{Memory.}
Compared to other methods, SLDA is extremely memory efficient. SLDA requires only 0.001 GB of storage for its covariance matrix, making it well-suited for memory constrained devices.  Conventionally, End-to-End and iCaRL store 20 images per class for ImageNet, which requires 3.011 GB of additional storage beyond the parameters of ResNet-18. ExStream stores 20 prototype vectors per class, requiring 0.041 GB of storage. 

\begin{figure}[t]
    \centering
    \includegraphics[width=0.35\textwidth]{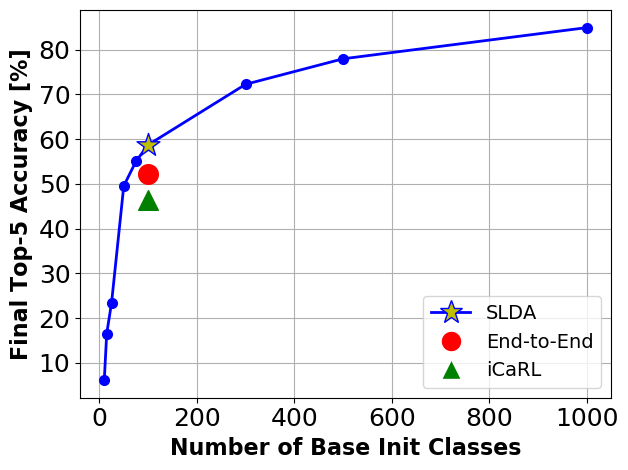}
    \caption{Final SLDA accuracy on ImageNet as a function of the number of base initialization classes. We denote performance of three models at 100 classes, which is the common initialization approach for ImageNet~\cite{castro2018end,rebuffi2016icarl}.}
    \label{fig:perf-vs-base-init}
    \vspace{-3mm}
\end{figure}

\input{tables/domain-transfer-results.tex}

\paragraph{Base Initialization.}
SLDA is reliant on robust deep feature representations in $G(\cdot)$ to achieve high classification accuracies. These features are thus dependent on the number of classes included in the base initialization phase and in Fig.~\ref{fig:perf-vs-base-init}, we plot the final top-5 performance of SLDA on ImageNet as a function of the number of classes used for this initialization. While using 100 classes for initialization with ImageNet is the standard approach~\cite{castro2018end,rebuffi2016icarl}, we find that the representations learned from only 50 classes provide SLDA with robust enough features to outperform iCaRL and using 75 classes allows SLDA to outperform both iCaRL and End-to-End. These results suggest consideration of whether incremental representation learning improves performance.

\paragraph{Domain Transfer.}
Since we use features from pre-initialized models for our experiments, we were interested in examining how SLDA behaves when the CNN features are initialized on different datasets. Specifically, we were interested in how well SLDA would perform when $G(\cdot)$ was initialized on ImageNet directly and $F(\cdot)$ was trained on CORe50 in the streaming setting, i.e., base initialization is performed only using the ImageNet dataset and does not include any data from CORe50. We conducted four variants of the experiment: 1)  $G(\cdot)$ was initialized on ImageNet and $\Sigma$ was initialized to a matrix of ones, 2) both $G(\cdot)$ and $\Sigma$ were initialized on ImageNet, 3) $G(\cdot)$ was initialized on ImageNet and then fine-tuned on the first 1,200 samples of CORe50 and $\Sigma$ was initialized to a matrix of ones, and 4) $G(\cdot)$ was initialized on ImageNet and then fine-tuned on the first 1,200 samples of CORe50 and $\Sigma$ was initialized on the first 1,200 samples of CORe50, which is consistent with our main experiments. 

Results for this experiment are provided in Table~\ref{tab:domain-transfer}. These results demonstrate that initializing $G(\cdot)$ and $\Sigma$ on data from CORe50 yielded the best results for the iid and instance orderings, but initializing only $G(\cdot)$ on CORe50 yielded the best results for both class orderings. However, when $G(\cdot)$ was initialized on ImageNet and the covariance was plastic, the largest difference in performance from the CORe50 initialized model was 5.6\% for the iid ordering and the smallest difference was only 1.0\% for the class iid ordering. Interestingly, the SLDA model with a covariance matrix initialized to ones performed the best for the class orderings. We hypothesize that this is because the covariance matrix does not overfit to the base classes, as is the case when the model performs a base initialization phase with the first two classes of CORe50. Although performing a base initialization phase with CORe50 often yielded higher results, Table~\ref{tab:domain-transfer} suggests that SLDA is capable of domain transfer from ImageNet to CORe50, without requiring a base initialization phase. This makes SLDA more amenable to applications where a user already has good feature representations and would like to immediately begin streaming learning on a different dataset.


\section{Discussion \& Conclusion}
\label{sec:discussion}

Although SLDA is popular in the data mining community, it has not recently been used for streaming learning on large classification datasets. We revisited SLDA and combined it with a CNN. While our approach is simple, it is extremely effective, exceeding recent incremental batch learning methods that loop through the dataset, while being much more lightweight. SLDA mitigates catastrophic forgetting even under different data orderings, demonstrating its robustness and utility for non-stationary data streams that are more realistic than iid data streams. Despite only training the output layer of the network, SLDA outperforms iCaRL by 6\% and over 11\% in terms of $\Omega_{\mathrm{all}}$ on ImageNet and CORe50 respectively. This result is impressive since iCaRL requires updating the entire network, which uses more computational time and resources. While our offline results indicate greater performance is achievable by training the hidden layers after base initialization, we urge developers of future incremental learning algorithms to test simply training the output layer after base initialization to ensure gains are being realized. 

While we initialized SLDA using the standard base initialization procedure used by iCaRL and others, the covariance matrix could instead be initialized using large amounts of unlabeled imagery (i.e., self-taught learning~\cite{raina2007self}). This approach could be used to initialize a model with a good representation before streaming learning occurs. 

If compute and memory are not significant factors, an interesting future direction would be to combine SLDA with a rehearsal-based scheme. SLDA could be used for rapid learning, while key observations are stored for rehearsal. Rehearsal could occur when the agent is inactive to update the entire CNN, rather than only the output layer. The challenges would be determining whether to use the main network output layer or the SLDA model and how to handle feature drift. Similar to \cite{parisi2018lifelong}, another future direction could include the creation of an SLDA model that accounts for the temporal structure of video data in its update and inference procedures.


\appendix
\section*{Appendix}
\label{sec: appendix}
\counterwithin{figure}{section}
\counterwithin{table}{section}
\counterwithin{subsection}{section}
\renewcommand{\thefigure}{A\arabic{figure}}
\renewcommand{\thetable}{A\arabic{table}}
\renewcommand{\thesubsection}{A.\arabic{subsection}}
\setcounter{figure}{0} 
\setcounter{table}{0}

The offline model uses SGD with momentum=0.9 and weight decay=1e-4. For ImageNet, we use 90 epochs (batch size=128) and learning rate of 0.1 decayed by a factor of 10 at 30 and 60 epochs. We use standard data augmentation of random flips and random resized crops at 224$\times$224 pixels. For CORe50, we use 40 epochs (batch size=256) and learning rate of 0.01 decayed by a factor of 10 at 15 and 30 epochs. For ImageNet, we use the iCaRL parameters from \cite{rebuffi2016icarl}, and for CORe50, we use: exemplars=20 per class, epochs=60, weight decay=1e-4, batch size=64, learning rate=0.01 decayed by a factor of 5 at 20 and 40 epochs. For ExStream, we use the offline model parameters and follow iCaRL by storing 20 exemplars per class. We use the regularization model implementations/parameters from \cite{chaudhry2019efficient}.


\ifthenelse{\boolean{ack}}{
\paragraph*{Acknowledgments.} This work was supported in part by NSF award \#1909696, the DARPA/MTO Lifelong Learning Machines program [W911NF-18-2-0263], and AFOSR grant [FA9550-18-1-0121]. The views and conclusions contained herein are those of the authors and should not be interpreted as representing the official policies or endorsements of any sponsor. We also thank fellow lab members Robik Shrestha and Ryne Roady for their comments and useful discussions.
}


{\small
\bibliographystyle{ieee_fullname}
\bibliography{egbib}
}

\end{document}

%% file: tables/main-results-table.tex
\begin{table*}[th!]
\caption{$\Omega_{\mathrm{all}}$ classification results on ImageNet and CORe50. We specify the plastic/updated (plas.) parameters and streaming (str.) methods. For CORe50, we explore performance across four different ordering schemes and report the average over 10 runs. All models use ResNet-18. The best \emph{streaming} model for each dataset and ordering is highlighted in \textbf{bold}.}
\label{tab:classification-results}
\centering
\begin{tabular}{lccccccc}
\toprule
& \multirow{3}{*}{\textsc{Plas.}} &  \multirow{3}{*}{\textsc{Str.}}  & \textsc{\textbf{ImageNet}} & \multicolumn{4}{c}{\textsc{\textbf{CORe50}}} \\ 
 \cmidrule(r){4-4} \cmidrule(r){5-8}
 \textsc{Ordering Scheme} &  &  & \textsc{cls iid} & \textsc{iid} & \textsc{cls iid} & \textsc{inst} & \textsc{cls inst} \\ 
\midrule
\emph{Output Layer Only} &  & & & & & & \\
Fine-Tuning & $\theta_F$  & Yes & 0.146 & 0.975 & 0.340 & 0.916 & 0.341 \\
ExStream~\cite{hayes2019memory} & $\theta_F$  & Yes & 0.569 & 0.953 & 0.873 & 0.933 & 0.854 \\
SLDA (Fixed $\Sigma$) & $\theta_F$ & Yes & 0.748 & 0.967 & 0.916 & 0.943 & 0.913 \\
SLDA (Plastic $\Sigma$) & $\theta_F$  & Yes & \textbf{0.752} & \textbf{0.976} & \textbf{0.958} & \textbf{0.963} & \textbf{0.959} \\
\midrule
\emph{Representation Learning} & &  & & & & & \\
Fine-Tuning & $\theta_F, \theta_G$ &  Yes & 0.121 & 0.923 & 0.334 & 0.287 & 0.334 \\
iCaRL~\cite{rebuffi2016icarl} & $\theta_F, \theta_G$ &  No & 0.692 & - & 0.839 & - & 0.845 \\
End-to-End~\cite{castro2018end} & $\theta_F, \theta_G$ &  No & 0.780 & - & - & - & - \\

\midrule \midrule
\emph{Approximate Upper Bounds} &  & & & & & & \\
Offline (Last layer) & $\theta_F$  & No & 0.853 & 0.979 & 0.954 & 0.966 & 0.955 \\
Offline  & $\theta_F, \theta_G$  & No & 1.000 & 1.000 & 1.000 & 1.000 & 1.000 \\
\bottomrule
\end{tabular}
\vspace{-4mm}
\end{table*}

%% file: tables/regularization-table.tex
\begin{table}[t]
\caption{$\Omega_{\mathrm{all}}$ results for regularization models averaged over 10 runs on CORe50 with and without Task Labels (TL). \label{table:regularization-results}}
\centering
\footnotesize
\begin{tabular}{lcccc}
\toprule
& \multicolumn{2}{c}{\textsc{\textbf{cls iid}}} & \multicolumn{2}{c}{\textsc{\textbf{cls inst}}} \\
\cmidrule(r){2-5}
\textsc{Model} & \textsc{TL} & \textsc{No TL} & \textsc{TL} & \textsc{No TL} \\ 
\midrule
SI~\cite{zenke2017continual} & 0.895 & 0.417 & 0.905 & 0.416 \\
EWC~\cite{kirkpatrick2017} & 0.893 & 0.413 & 0.903 & 0.413 \\
MAS~\cite{aljundi2018memory} & 0.897 & 0.415 & 0.905 & 0.421 \\
RWALK~\cite{chaudhry2018riemannian} & 0.903 & 0.410 & 0.912 & 0.417 \\
A-GEM~\cite{chaudhry2019efficient} & 0.925 & 0.417 & 0.916 & 0.421 \\
SLDA (Fixed $\Sigma$) & 0.973 & 0.916 & 0.971 & 0.913 \\
SLDA (Plastic $\Sigma$) & \textbf{0.989} & \textbf{0.958} & \textbf{0.987} & \textbf{0.959} \\
\midrule
Offline & 1.000 & 1.000 & 1.000 & 1.000 \\
\bottomrule
\end{tabular}
\vspace{-4mm}
\end{table}

%% file: tables/domain-transfer-results.tex
\begin{table*}[t]
\caption{$\Omega_{\mathrm{all}}$ classification results for the domain transfer experiment from ImageNet to CORe50 with SLDA using a fixed and plastic (plas.) covariance matrix ($\Sigma$). We indicate the dataset used for base initialization of the SLDA parameters ($G$ and $\Sigma$).}
\label{tab:domain-transfer}
\centering
\begin{tabular}{lcccccccc}
\toprule
& \multicolumn{2}{c}{\textbf{\textsc{iid}}} &  \multicolumn{2}{c}{\textbf{\textsc{cls iid}}}  & \multicolumn{2}{c}{\textbf{\textsc{inst}}} &  \multicolumn{2}{c}{\textbf{\textsc{cls inst}}} \\ 
 \cmidrule(r){2-9}
\textsc{Base Init.} & \textsc{Fixed} & \textsc{Plas.} & \textsc{Fixed} & \textsc{Plas.} & \textsc{Fixed} & \textsc{Plas.} & \textsc{Fixed} & \textsc{Plas.} \\ 
 \midrule 
ImageNet ($G$) & 0.868 & 0.920 & 0.926 & 0.948 & 0.861 & 0.924 & 0.927 & 0.948 \\
ImageNet ($G$, $\Sigma$) & 0.860 & 0.888 & 0.909 & 0.934 & 0.856 & 0.892 & 0.909 & 0.934 \\
CORe50 ($G$) & 0.963 & 0.968 & \textbf{0.967} & \textbf{0.970} & 0.936 & 0.947 & \textbf{0.963} & \textbf{0.971} \\
CORe50 ($G$, $\Sigma$) & \textbf{0.967} & \textbf{0.976} & 0.916 & 0.958 & \textbf{0.943} & \textbf{0.963} & 0.913 & 0.959 \\
\bottomrule
\end{tabular}
\vspace{-3mm}
\end{table*}